# Enhancing the Capabilities of Large Language Models for API calls through Knowledge Graphs

Ye Yang [a,*], Xue Xiao[a], Ping Yin[a], Taotao Xie[a]

a, INSPUR GROUP CO., LTD.,250101 Jinan City, Shandong Province,China

13054558662@163.com

Corresponding author at: INSPUR GROUP CO.,LTD.，No. 1036, Langchao Road, High-tech Zone, 250101 Jinan City, Shandong Province, China.

E-mail address:13054558662@163.com(Yang Ye).

## Abstract

API calls by large language model (LLM) represent a cutting-edge technique in data analysis. However, the potential of LLM to effectively utilize tools through API calls remains underexplored in knowledge-intensive sectors such as the meteorological industry. In this paper, we propose a system, named KG2data, that integrates knowledge graphs, LLM, React agents, and tool usage technologies to perform API calls for intelligent data acquisition and query handling in the meteorological domain. We test the accuracy of the system's API calls using a virtual API. The baseline systems for comparison are chat2data (KG2data without knowledge) and RAG2data (KG2data with a vector database replacing the knowledge graph). Our experimental results demonstrate that the proposed system (1.43%, 0% and 88.57% in 3 evaluation metrics) outperforms RAG2data (16%, 10% and 72.14% in 3 evaluation metrics) and chat2data (7.14%, 8.57% and 71.43% in 3 evaluation metrics) in terms of failure rate of name recognition, failure rate of hallucination recognition and accuracy rate for API calls. Our system integrates knowledge graph, LLM, and ReAct-master Agent technologies. Unlike current LLM used for API calls, our system overcomes the challenge of limited domain-specific knowledge of LLM, which often makes it difficult to address complex queries containing specialized terminology or lengthy questions. By utilizing knowledge graphs as long-term memory, our system significantly improves conten-retrieval coverage, handling of complex queries, industry-specific logical reasoning, deep semantic relationships among entities, and the integration of heterogeneous data. Additionally, it addresses the high computational costs associated with training or fine-tuning LLM, making it more adaptable to the dynamic nature of domain knowledge and APIs. In summary, the KG2Data system offers a fresh perspective for intelligent knowledge-based question answering and data analysis in knowledge-intensive industries.

# 1. Introduction

With the increasing frequency of climate change and extreme weather events, various industries are facing far-reaching impacts. Drought, flood and high temperature not only lead to crop reduction, threaten food security, but also affect the stability and efficiency of hydropower generation. In addition, floods, storms and high temperatures may damage roads, bridges and railways, thus affecting transportation. Frequent extreme weather events also increase the pressure of emergency response, making meteorological disaster prevention and mitigation the focus of great attention of governments at all levels and all sectors of society **(AghaKouchak, Chiang et al. 2020, Ebi, Vanos et al. 2021, Sindall, Mecrow et al. 2022)**. In order to effectively address these challenges, accurate meteorological services have become crucial, which poses challenges to the processing and application of large-scale meteorological data. With the rapid development of new generation information technologies such as large artificial intelligence, cloud computing and big data, the original application support capacity of meteorological big data has been difficult to meet the growing intelligent needs **(Yadav and Chandel 2014, Cabaneros, Calautit et al. 2019, Slater, Arnal et al. 2023)**. The traditional form of meteorological data utilization has significant limitations, mainly manifested in the following aspects:

Difficulty in effective utilization of data: Meteorological data is highly diverse and complex, and traditional data modeling methods are difficult to process and analyze these data efficiently; it is difficult to sort out the data relationships and it depends on human experience, further increasing the operational difficulty (Sun, Deng et al. 2024).

Data confidentiality issue: Meteorological data involves national security, military use and economic interests, and sensitive data restricts public sharing and commercial application (Qin Yunlong, Wang Yingying et al. 2020).

Data valuation problem: The large-scale demand, high-value application and technological accumulation of meteorological data in China provide good market and supply conditions for meteorological data services to release their high value.

However, the value of meteorological data elements has not formed a standardized and normalized system of value evaluation**(Belward and Skoien 2015, Frazier and Hemingway 2021)**.

Challenges in Addressing the Diverse Needs of Multi-Industry Applications: Commercial meteorological services have reached a mature stage in Europe and the United States, generating annual revenues in the hundreds of billions of U.S. dollars. In contrast, this industry in China remains in its nascent phase, yet it holds significant growth potential. The vast scale and diversity of meteorological data present considerable challenges in data retrieval, particularly for non-expert users. As a result, the accessibility and usability of such data are currently more suited for professionals with specialized knowledge in the meteorological field.

Over the past decade, machine learning, such as Learning-driven optimizer (Yu, Chai et al. 2022), learned index (Sun, Zhou et al. 2023), and Intelligent tuning of database parameter configuration (Zhou, Li et al. 2023), have been widely applied across various industries to enhance data systems, aiming to reduce user barriers and improve efficiency. However, these methods exhibit a high dependency on training data, limited generalization capabilities, and challenges in adapting to real-time adjustments in data systems as they evolve over time. Since 2022, the gradual maturation of generative large language model (LLM), such as GPT, has demonstrated impressive performance in understanding user intent and generating task-specific outputs. In particular, these models have shown significant potential in optimizing data systems and improving data governance(Tang, Fan et al. 2021, Zhao, Lim et al. 2023, Zhou, Sun et al. 2024).

Despite the superior performance of LLM in many fields, challenges remain in leveraging these models for automated, intelligent, and precise data querying, recommendation, and user-specific analysis. Key challenges include hallucinations (Sallam 2023, Samsi, Zhao et al. 2023) due to the lack of domain-specific knowledge, high inference costs associated with LLM, and inaccurate reasoning results for tasks requiring high precision(Sallam 2023).

To address these challenges, some studies have enabled large language models (LLM) to utilize external tools (Schick, Dwivedi-Yu et al. 2024), allowing them to access larger and dynamically evolving knowledge bases and perform a wide range of subtle tasks. Building upon this, the interaction between LLM and external APIs has gradually become a research hotspot. By providing access to computational tools, research by (Thoppilan, De Freitas et al. 2022) has demonstrated that enhanced LLM can handle larger, more dynamic knowledge spaces and perform complex computational tasks. Consequently, leading LLM providers(OpenAi, Achiam et al. 2023)have begun integrating plugins to enable these models to call external tools via application programming interfaces (APIs). This integration allows users to invoke complex software functions through simple inputs, thereby improving interaction efficiency and lowering the barriers to software usage.

However, many previous studies integrating APIs into large language models(Liang, Wu et al. 2024) have seldom considered the system's performance in specialized domains, such as meteorology. In fact, in the context of domain-specific tasks, user queries are often semantically implicit and contain substantial background information. Such semantically ambiguous queries may lead large language models (LLM) to hallucinate responses. Moreover, due to the lack of domain expertise, LLM may struggle to fully understand user intentions within specialized fields. Lastly, domain-specific knowledge is inherently dynamic, and LLM, due to the high computational cost of retraining and fine-tuning, typically find it difficult to adapt to the evolving nature of both domain knowledge and APIs. These three challenges make the standard approach of using LLM for API calls difficult to adapt to specialized domains.

In this context, (Zhao, Zhou et al. 2024) have innovatively proposed a conversational, interactive data analysis platform driven by Large Language Models (LLM) and Retrieval-Augmented Generation (RAG) techniques, referred to as

RAG2Data. RAG2Data leverages RAG technology to manage unstructured data within specialized domains, while structured data is retrieved via a Text-to-SQL approach. This design significantly reduces the need for direct interaction with LLM, mitigating issues such as hallucinations, low reasoning accuracy, and high inference costs commonly associated with LLM. However, the vector similarity search mechanism of RAG presents several challenges in current applications: (1) the retrieval coverage is limited, making it difficult to uncover deep semantic relationships between data; (2) it struggles to perform complex queries; (3) it cannot effectively integrate heterogeneous data from diverse sources; (4) it fails to perform reasoning and extension of knowledge in specialized domains; and (5) it suffers from delays in data updates.

Recent studies have focused on combining Large Language Models (LLM) with knowledge graphs to build interactive question-answering knowledge bases tailored to specialized domains, aiming to address the limitations of Retrieval-Augmented Generation (RAG) technology **(Arsenyan, Bughdaryan et al. 2023, Oladeji, Mousavi et al. 2023, Soman, Rose et al. 2024)**. A knowledge graph serves as a structured framework for representing knowledge relationships. Its structured associative architecture allows for the integration of datasets from diverse sources **((Lairgi, Moncla et al. 2024, Sun, Luo et al. 2024)**. This facilitates deep semantic relationship mining (Wang, Chen et al. 2024) and knowledge reasoning extension within specialized domains 推理扩展(Lan, He et al. 2021, Zhu, Zhang et al. 2022). Moreover, knowledge graphs support complex queries, such as path and subgraph queries, and have shown strong performance in knowledge-intensive tasks(Schneider, Klettner et al. 2024). These advantages are not inherent to RAG technology. Consequently, leveraging knowledge graphs to enhance LLM performance in specialized tasks has become a significant trend in current development.

Currently, few technologies leverage knowledge graphs to enhance the performance of large language models (LLM) in specialized tasks, such as the precise retrieval, analysis, and processing of data within the context of expert knowledge semantics. To address the following challenges: (1) the inefficiency in meteorological data acquisition and utilization, as well as data confidentiality issues, (2) the limitations of Retrieval-Augmented Generation (RAG) technology in performing complex queries and its lack of knowledge reasoning capabilities, and (3) the absence of domain-specific expertise in current LLM-based API calling methods, our team proposes an innovative system, **LLM-Driven Meteorological KG2Data**. This system integrates knowledge graph technology, React-based expert agent technology, LLM, and API calls to achieve intelligent, automated meteorological data acquisition and analysis within a legal and compliant framework.

## 2. Related Work

## 2.1 LLM for Tool Usage

Large Language Models (LLM) have made significant advancements in the field of natural language processing. The concept of LLM for Tool Usage refers to leveraging LLM to interact and collaborate with various tools to accomplish more complex tasks and solve real-world problems. These tools can include, but are not limited to, the following types:

External Databases and Knowledge Graphs: LLM can access external databases and knowledge graphs to obtain more accurate and comprehensive information, thereby enhancing the quality of responses and text generation. For example, when addressing domain-specific queries, LLM can utilize specialized databases to retrieve the latest research findings and data (Huang, Parthasarathi et al. 2024).

Software Tools and APIs: Integration with various software tools and application programming interfaces (APIs) enables LLM to perform specific and subtle tasks, such as data analysis and image recognition. For instance, by invoking mathematical function libraries, LLM can carry out complex mathematical computations (Zhang, Chen et al. 2023).

Agents and Automation Systems: As agents, LLM can collaborate with other automation systems to execute tasks autonomously and optimize workflows. For example, in software development, LLM can interact with agents and tools to enhance development efficiency (Tufano, Mastropaolo et al. 2024).

## 2.2 LLM for API calls

API calls have become a cutting-edge focus in the application of LLM, with increasing encouragement to use them as tools (Patil, Zhang et al. 2023, Shen, Song et al. 2024). This technology enables users to retrieve and process data through natural language queries, eliminating the need for complex programming knowledge. By transforming natural language into a JSON format that meets the specific requirements of APIs, the data analysis process becomes more intuitive and efficient (Escarda-Fernández, López-Riobóo-Botana et al. 2024). However, limited research has been conducted on API calls in vertical domains, which are crucial for the precise acquisition and analysis of specialized data.

## 2.3 RAG2data

RAG2data, proposed by (Zhao, Zhou et al. 2024), is a data analysis platform enhanced by Large Language Models (LLM) and Retrieval-Augmented Generation (RAG) techniques. The architecture of RAG2data comprises three layers. The knowledge management layer is responsible for collecting and preprocessing data, splitting it, selecting embedding models, storing it in a vector database, and managing tools. The online query inference layer handles query preprocessing, converts queries into vectors, analyzes intent, and either retrieves knowledge and APIs for single-round LLM input or utilizes an LLM agent for multi-round pipeline generation. This layer also leverages the vector database for caching purposes. The LLM layer generates results, summarizing unstructured data, translating natural language into SQL, and employing pandas APIs for structured data analysis and visualization.

Previous research has largely overlooked the performance of API calls in specialized domains, either due to challenges in data acquisition within these fields or a lack of knowledge inference capabilities specific to the domain. Additionally, the high cost of training and fine-tuning large language models (LLM) makes it difficult for them to adapt to the dynamic changes in APIs and domain-specific knowledge. In contrast, knowledge graphs offer a promising solution to address these limitations.

Unlike previous studies, our research focuses on a more constrained domain—API calls in specialized fields, exemplified by atmospheric science. By integrating knowledge graphs, agents, LLM, and tool usage, we enable API calls that enhance the performance of LLM in domain-specific data acquisition and analysis, without the need for complex fine-tuning or dealing with low-level implementation details.

# 3. Metohdology

In this section, we describe the construction process of KG2data and the associated technologies. First, we outline the process of preparing the API dataset and the creation of instruction-answer pairs used for technical performance evaluation. Subsequently, we introduce KG2data, an innovative framework driven by Large Language Models (LLM) that integrates domain-specific knowledge graphs to facilitate domain-specific API calls. Finally, we present the matching evaluation metric employed to assess the system's performance in executing API calls.

## 3.1 Dataset Collection

### 3.1.1 API Documentation

Due to data security concerns, many meteorological APIs are not freely accessible. **(Liu, Pei et al. 2023)** propose the generation of virtual APIs to evaluate model performance on tasks for API calls. Based on this approach, we reviewed the documentation of APIs and datasets from leading organizations and websites. Using this information, we designed 30 virtual APIs related to the meteorological domain (as shown in Figure 1). These APIs cover common meteorological parameters in atmospheric science, such as temperature, humidity, precipitation, wind speed, wind direction, atmospheric pressure, and radiation. Both input and output parameters are formatted in JSON.

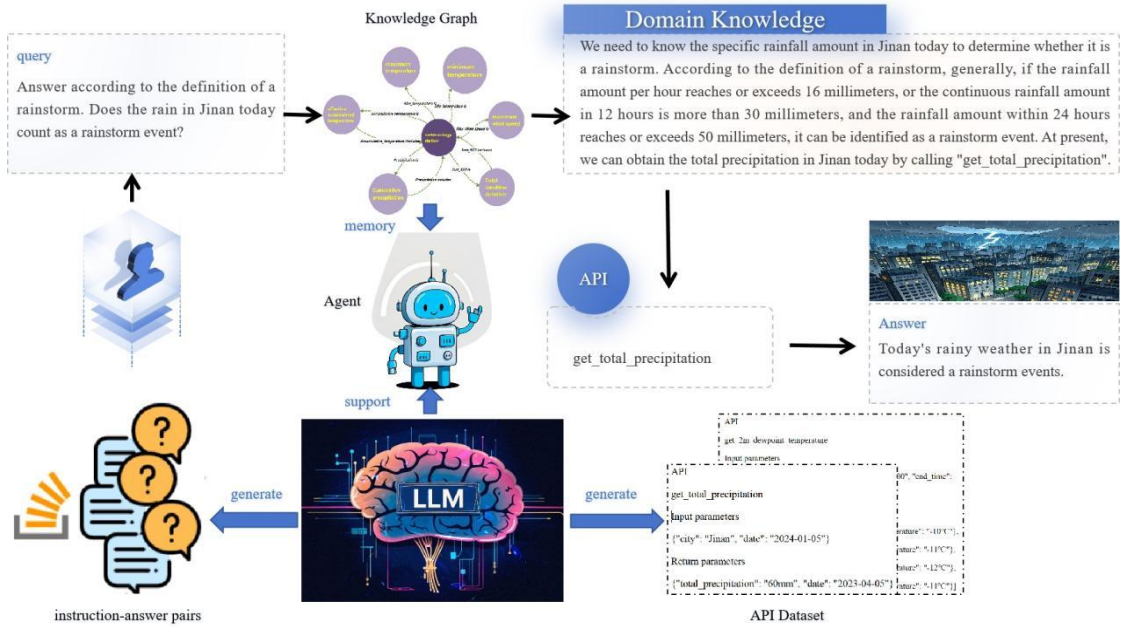

Figure 1: KG2data: an interactive data analysis system that integrates a knowledge graph, an agent, and a large language model (LLM) to facilitate interactions with API. The lower half of the system illustrates the preliminary preparation of API datasets and instruction-answer pairs, as outlined in Section 3.1.The upper half depicts the architecture and operational process of KG2data, as detailed in Section 3.2. In this example, the system is capable of suggesting the appropriate API calsl to generate answers from a user's natural language query.

### 3.1.2 Instruction-Answer Pairs

We require Large Language Models (LLM) to utilize the self-instruct paradigm (Wang, Kordi et al. 2022) to generate instruction-answer pairs for weather-related tasks for API calls . By providing contextual examples and API documentation with reference information, we specifically instruct LLM to avoid using any API names or explicit prompts when generating instructions (see Figure 1). For each API, we construct two instruction-answer pairs: one with a clear question and another with a more specialized, implicitly framed question. This design aims to evaluate the system's ability to perform API calls under both simple and domain-specific inquiry scenarios. This is crucial, as highly specialized phrasing in questions posed by domain experts may be too complex for LLM to comprehend, potentially impacting the API calls.

## 3.2 KG2data

### 3.2.1 System Overview

We propose an LLM-driven KG2data framework to enhance the capability of API calls in the field of atmospheric science. The system consists of four key components to facilitate API calls, including a meteorological knowledge graph, an agent, a Large Language Model (LLM), and a calling tool. An overview of our system is depicted in Figure 1. Compared to traditional API calls, our system utilizes a domain-specific knowledge graph as its memory module, allowing the agent to search for APIs that meet specific task requirements, providing reliable answers enriched with professional knowledge. In the following sections, we will present the four components of our framework in detail.

### 3.2.2 Knowledge Graph

In the meteorological scenario, meteorological processes involve both unstructured textual descriptions and structured data support. The unstructured text corresponds to meteorological domain knowledge, such as the description of a heavy rainfall event, while the structured data refers to values of specific meteorological element, such as precipitation levels during a particular period of the rainfall event. Therefore, it is essential to construct a joint knowledge graph that integrates meteorological terminology and meteorological datasets.

This knowledge graph is a multi-dimensional, multi-layered integrated dataset focused on the meteorological domain. It uses large language models for entity recognition, relationship extraction, and community summarization algorithms to construct both an empirical and a data-driven knowledge graph for meteorology. The approach involves establishing multi-level semantic associations and graph aggregation between the knowledge graph of meteorological data and the meteorological expertise . Additionally, it analyzes the community structure and redundancy distribution within the meteorological knowledge graph, optimizing and pruning the graph by incorporating expert knowledge in a human-machine collaboration model. The specific methodology is outlined as follows:

This study utilizes large language models (LLM) to automatically chunk corpus data in meteorological domain, exploring token specifications that optimize the balance between precision and recall in retrieval of knowledge base . Leveraging the in-context learning capabilities of LLM, we develop small-sample prompt templates for identifying attributes of data element and extracting relationships specific to meteorological scenarios. The LLM are then employed to extract features and relationships of entity from the corpus, with a focus on knowledge graph in the meteorological domain. Additionally, the Leiden algorithm is applied to discover the multi-level community within the knowledge graph and to prune redundant relationships, thereby reducing retrieval redundancy.

The knowledge graph fully harnesses the transparent interpretability of knowledge graphs, using their knowledge navigation and integration capabilities to enhance the reasoning capabilities of system within specialized domains.

The ReAct-Expert Agent is built upon an enhanced knowledge graph, designed to support the querying and analysis of meteorological data. It provides personalized recommendations of meteorological data, tailored to the needs of specific users. Additionally, the system offers a service for retrieving and orchestrating behavior-driven scenario data, specifically for applications related to meteorological event and process analysis.

### 3.2.3 ReAct-expert Agent

The ReAct-expert agent ensures that the results are a query-focused summarization (QFS) of meteorological data and weather process in question. The process involves the expert agent feeding a combination of knowledge graphs, prompt, and tools results into the large language model (LLM). The agent is responsible for retrieving meteorological data and document from the API using tools, processing it as needed, and evaluating the outcomes to ensure accuracy. The knowledge graph as the memory module of the agent guarantees that the output is highly relevant to the weather process. This enables the agent to perform Long-Range Correlation with meteorological knowledge and automatically call multiple tools.

Especially, the expert agent uses the ReAct framework, a general-purpose paradigm that integrates reasoning and action with LLM (Yao, Zhao et al., 2022). In this approach, LLM generate reasoning trajectories and task-specific actions in an interleaved manner. The reasoning trajectories allow the model to generate, track, and update action plans, even addressing exceptional cases. The action steps facilitate interaction with external sources, such as knowledge bases or environments, to gather information and provide more reliable and actionable responses. The ReAct agent outperforms several state-of-the-art agent baselines in both language and decision-making tasks. Furthermore, ReAct enhances the interpretability and trustworthiness of LLM. In this study, the ReAct agent breaks down the reasoning chain into the steps shown in Figure 2.

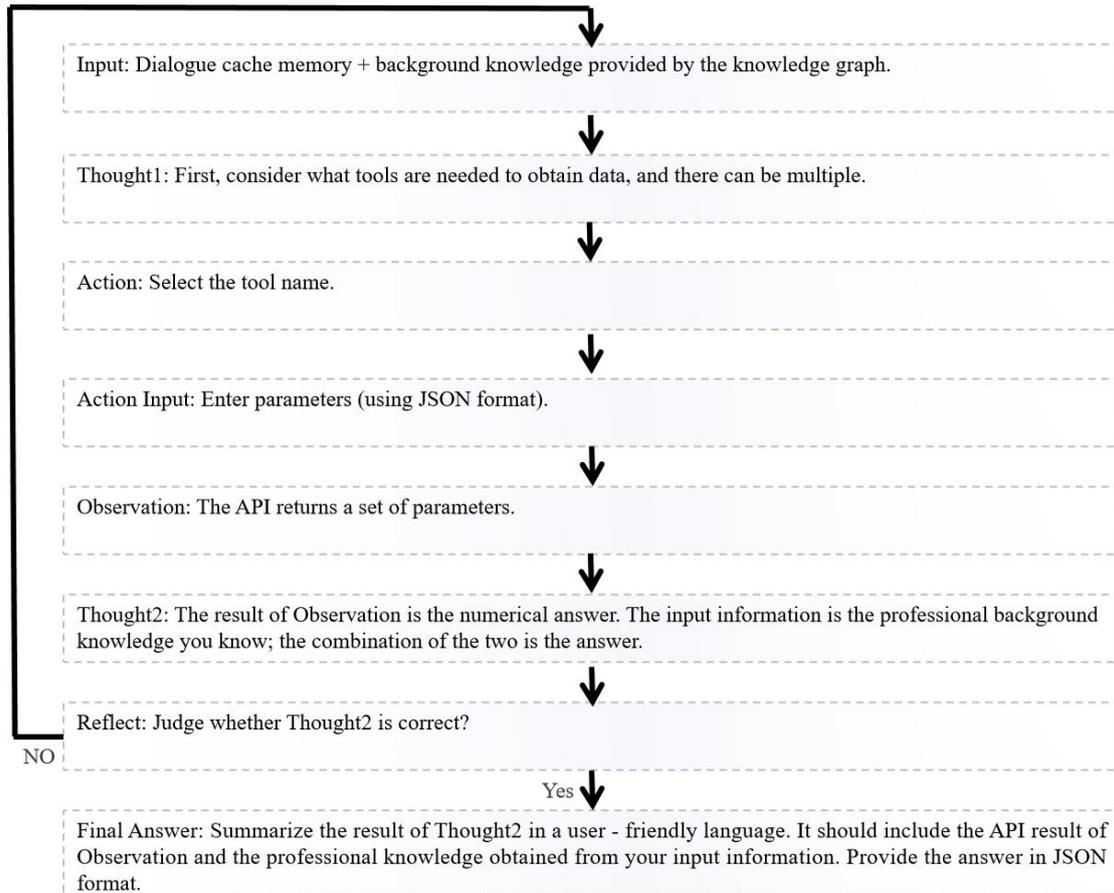

Figure2: The reasoning chain of ReAct agent in this paper.

### 3.2.4 Tools for Data Acquisition

The data acquisition tools connect APIs with ReAct expert agents, enabling data collection and analysis based on the needs of each meteorological scenario. Each tool is characterized by a unique name, parameters, prompt words, and information related to the integrated API.

These tools are designed to extract relevant fields from the returned result, perform statistical analysis and processing, and then deliver the results to the user through a conversational interface.

**Definition of Data Tool**:

$$Data\ Tool \rightarrow Parameter\ Extraction \rightarrow API\ Calls \rightarrow Data\ Return$$

The names of the data tools should be designed to be easily understood by large language model (LLM) agents for intent recognition. The parameters of the tools should be structured to allow for straightforward extraction through conversational queries, enabling the LLM agents to identify the necessary inputs effectively. Additionally, the descriptions of the tools should be phrased in a way that improves the LLM accuracy in recognizing intent, thereby reducing the likelihood of misclassification or incorrect tool usage.

## 3.3 Verifying APIs

We assess the performance of the system's API calls using the metrics proposed by (Liu, Pei et al. 2023), including API-calls accuracy rate (ACAR), failure rate of intent recognition (FRIR), failure rate of name recognition (FRNR), failure rate of parameter recognition (FRPR), and failure rate of hallucination recognition of API (FRHR). These metrics are defined as follows:

$$ACAR = \frac{Number\ of\ correct\ API\ calls}{Total\ number\ of\ API\ calls}$$

$$FRIR = \frac{Number\ of\ failure\ rate\ of\ intent\ recognitions\ for\ API\ calls}{Total\ number\ of\ API\ calls}$$

$$FRNR = \frac{Number\ of\ failure\ rate\ of\ name\ recognition\ for\ API}{Total\ number\ of\ API\ calls}$$

$$FRPR = \frac{Number\ of\ failure\ rate\ of\ parameter\ recognition\ for\ API\ calls}{Total\ number\ of\ API\ calls}$$

$$FRHR = \frac{Number\ of\ failure\ rate\ of\ hallucination\ recognition\ for\ API}{Total\ number\ of\ API\ calls}$$

To track which API is called by the LLM, we visualize the reasoning chain of the ReAct-master agent. By examining whether the API called in the [Observations] step is valid, we can determine whether the agent correctly performs the API call, which is used to calculate ACAR. Identifying and defining the "hallucination" phenomena in LLM presents a significant challenge. We detect hallucinations by checking whether the API invoked in the [Action] step is a fictitious one, which is used to calculate the FRHR for tools. It is important to note that "hallucination" differs from an incorrect API call: it refers to the invocation of a real API that does not address the specific question-answer task, or the failure to invoke any API at all (used to compute the FRNR). The [Action Input] step provides the parameters for the tools, which are used to calculate the FRPR. [Thought1] step analyzes the user's query intent and how background knowledge can assist in solving the problem, which is used to calculate the FRIR.

# 4 Experimental Evaluation

## 4.1 Baseline

We compare KG2data with RAG2data and CHAT2data. RAG2data replaces the knowledge graph module of the KG2data system with a vector database, which is constructed using the same data as KG2data. CHAT2data, on the other hand, removes the knowledge graph module from the KG2data system and only includes certain functional descriptions of the API within the prompt.

## 4.2 Accuracy on API Calls

In Table 1, we report the performance of our method alongside baseline methods in the context of meteorological API calls. It is evident that our method outperforms the baseline methods, demonstrating higher ACAR and lower rates of FRIR, FRNR, FPPR as well as FRHR.

**Table1: The comparison of the performance for KG2data, RAG2data, and Chat2data.** ** represents a significance level of 0.05, and * represents a significance level of 0.1. Row 4 indicates the significance level of KG2data relative to KAG2data, while Row 6 indicates the significance level of KG2data relative to chat2data.**

|  | FRIR | FRNR | FRPR | FRHR | ACAR |
|---|---|---|---|---|---|
| KG2data | 0.00% | 1.43% | 2.86% | 0.00% | 88.57% |
| RAG2data | 8.57% ** | 16% ** | 10.00% ** | 10.00% ** | 72.14% ** |
| chat2data | 1.43% | 7.14% * | 7.14% | 8.57% ** | 71.43% ** |

All the metrics of RAG2data are lower than those of KG2Data. Compared to RAG2data, KG2data enhances ACAR by 16.43%, reduces FRIR, FRNR,,FRPR, and FRHR with 8.57%, 14.57%,7.14% and 10% respectively. The knowledge graph equips KGdata with domain-specific meteorological reasoning capabilities, enabling it to effectively identify key information needed to answer user queries, even when these queries are implicit or lengthy, and may not directly contain API-related details (Example 1 in Figure3). Moreover, the triple-based knowledge representation efficiently filters out irrelevant, redundant information, retaining only the core entities and relationships essential to the query. The precise and specialized knowledge provided by the knowledge graph significantly improves the FRIR and ACAR in large language models (LLM) when performing API calls, while largely mitigating execution errors caused by LLM hallucinations (A lower FRHR of KG2data compared to RAG2data.). While RAG also provides background knowledge to RAG2data, this knowledge is derived through vector similarity-based retrieval from the vector database. In cases of implicit queries, RAG may fail to offer useful information. Furthermore, when queries are verbose, RAG typically retrieves large volumes of irrelevant information, which may lead to LLM hallucinations and hinder the system's ability to execute API calls effectively.

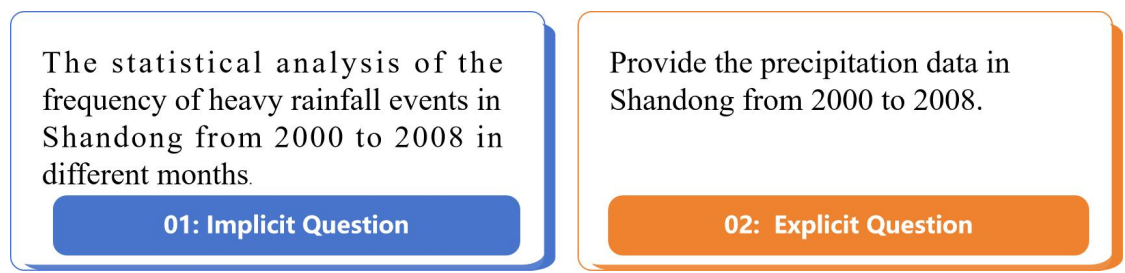

Figure 3: The implicit question and explicit question in meteorological domain.

KG2Data are significantly superior in with the FRNR, FRHR and ACAR of 5.71%, 8.57% and 17.24 in comparison to chat2data, respectively. Chat2data lacks domain-specific knowledge and exhibits insufficient reasoning ability. it is prone to causing hallucinations in large models or difficulty in correctly identifying the appropriate API, which making it difficult to perform API calls effectively.

Notably, FRIR, FRNR, FRPR, and FRHR are insufficient to explain the inaccuracy probability of KG2data in API call tasks. This is due to the fact that while the KG2Data system can accurately determine which API should be called during the

[observation] step of the reasoning chain of agent, the hallucinations of LLM result in the absence of return parameters of API calls in the final answer. This finding demonstrates that the performance of large models and the design of prompt templates significantly impact the final answer. In the future, we plan to implement more stringent prompt templates to enhance system performance in API calls. We also look forward to the release of more advanced LLM with improved capabilities.

# 5 Discussion

In this study, we propose an efficient framework that integrates a meteorological knowledge graph, ReAct-master agents, and LLM to address the challenges associated with the underutilization of meteorological datasets and the limitations of LLM-driven API calls in specialized domains due to knowledge gaps. The KG2data framework demonstrates superior performance in accuracy of API call. We compare the proposed framework with RAG2data and Chat2data. The results indicate that KG2data is more robust to prompt perturbations and achieves significantly higher in ACAR and lower in FRNR for API calls. Moreover, it reduces FRHR in API calls caused by LLM' hallucinations. We hypothesize that the improvement in performance is due to the fusion of explicit knowledge from the knowledge graph, sophisticated architectures from React-expert agents, and intent-recognition capabilities of the LLM. These findings highlight the value of providing fine-grained domain-specific knowledge at the prompt and emphasize the significance of designing reasoning-behavior-interleaved expert agents for guiding LLM in executing precise tasks.

It is important to note that knowledge graphs significantly improve the performance of LLM in domain-specific API call tasks. In this study, the integration of heterogeneous, domain-specific knowledge into the knowledge graph holds the potential to generate new insights by linking different entities (Baranzini, Borner et al. 2022). Knowledge-augmented LLM are able to produce reliable, comprehensive responses rooted in domain-specific expertise. Additionally, the reasoning capabilities of knowledge graphs in specialized fields enhance both the accuracy of LLM responses and their performance in handling complex queries. This forms the basis for enabling agents to accurately identify the most relevant APIs for user queries (Yasunaga, Ren et al. 2021). The structured framework of knowledge graphs also greatly strengthens the effectiveness of prompts (Pan, Razniewski et al. 2023), as the structured format allows for exponential growth in entity connections, making the injected knowledge far more information-dense than RAG.

In terms of architectural design, the KG2data system employs a modular, loosely coupled, and highly cohesive approach, dividing the system into distinct modules, each responsible for specific functions and tasks. This design philosophy enhances both scalability and maintainability.The KG2data system is built on a systems engineering framework, with a set of unified standards and specifications. It integrates a range of technologies, including data tools developed through domain-specific platform APIs, intelligent agent workflows, large model generation, and knowledge graphs. These integrations provide the system with high availability, smooth scalability, and rapid deployment capabilities.

In conclusion, our framework highlights the significance of combining LLM, knowledge graph, and ReAct-master expert agents, which greatly reduces the difficulty and cost for non-experts in accessing specialized data. However, due to time and resource limitations, the KG2data system has only been tested with 35 virtual APIs and 70 instruction-answer pair. Moving forward, we aim to explore a broader range of domain-specific API calls.

# 6 Conclusion

This study introduces the KG2data system, which enables the intelligent and automated retrieval of meteorological data through a conversational interface. The system overcomes the limitations of traditional API calls, which often lack domain-specific knowledge and tend to have low accuracy in specialized scenarios, such as data analysis. It also addresses the challenge of interpreting professional users' questions, which can be implicit or overly complex, making successful API calls difficult. The framework LLM, knowledge graph, ReAct-expert agent, and tool-utilization technologies. In this framework, the LLM (for reasoning and semantic understanding) complement the knowledge graph (which provides domain-specific knowledge, reasoning, and complex query capabilities), enhancing their combined effectiveness. The ReAct-master agent facilitates task decomposition and precise execution through an interwoven reasoning-action process. Specifically, the [Action] step identifies the API most relevant to the user's query, the [Action Input] step provides the necessary input parameters, and the [Observation] step presents the results of the API calls. This step-by-step breakdown effectively clarifies the reasoning process behind API calls in LLM, enabling easier identification of potential issues that could cause API calls failures.

Furthermore, this system addresses several limitations of RAG when applied to large-scale datasets, such as its inability to extract deep semantic associations, incomplete coverage in retrieval, difficulty in integrating data from multiple sources, delayed data updates, limited adaptability to complex query tasks, and the lack of reasoning and expansion for specialized domain knowledge. In contrast, knowledge graphs store knowledge in a graph structure, which clearly represents the complex relationships between entities, enabling LLM to comprehensively integrate and understand domain-specific knowledge. RAG, on the other hand, primarily relies on vector similarity for retrieval, making it less effective in uncovering deep semantic relationships. Knowledge graphs can integrate data from diverse sources, including structured, semi-structured, and unstructured data, into a unified knowledge graph. In contrast, RAG requires adaptation and integration of data from multiple sources, limiting its capability for complex queries. Knowledge graphs support complex queries based on graph structures, such as path queries and subgraph queries, facilitating various complex tasks in specialized domains. Conversely, RAG primarily supports keyword-based queries, limiting its ability to handle complex queries effectively. Additionally, knowledge graphs enable reasoning and expansion for domain-specific knowledge, while RAG lacks reasoning capabilities and mainly relies on the inherent reasoning abilities of LLM.

In conclusion, the KG2data framework we propose will significantly enhance the performance of LLM in specialized domain API applications. It also effectively addresses the challenges of high cost of training and fine-tuning for LLM, as well as their struggle to adapt to the dynamic changes in APIs and domain-specific knowledge. The KG2data framework shows great potential for improving the efficiency of data processing and utilization in specialized fields.

# 7 Reference

AghaKouchak, A., F. Chiang, L. S. Huning, C. A. Love, I. Mallakpour, O. Mazdiyasni, H. Moftakhari, S. M. Papalexiou, E. Ragno and M. Sadegh (2020). Climate Extremes and Compound Hazards in a Warming World. Annual Review of Earth and Planetary Sciences, Vol 48, 2020. R. Jeanloz and K. H. Freeman. **48:** 519-548.

Arsenyan, V., S. Bughdaryan, F. Shaya, K. Small and D. Shahnazaryan (2023). "Large Language Models for Biomedical Knowledge Graph Construction: Information extraction from EMR notes." Arxiv.

Baranzini, S. E., K. Borner, J. Morris, C. A. Nelson, K. Soman, E. Schleimer, M. Keiser, M. Musen, R. Pearce, T. Reza, B. Smith, B. W. Herr, II, B. Oskotsky, A. Rizk-Jackson, K. P. Rankin, S. J. Sanders, R. Bove, P. W. Rose, S. Israni and S. Huang (2022). "A biomedical open knowledge network harnesses the power of AI to understand deep human biology." Ai Magazine **43**(1): 46-58.

Belward, A. S. and J. O. Skoien (2015). "Who launched what, when and why; trends in global land-cover observation capacity from civilian earth observation satellites." Isprs Journal of Photogrammetry and Remote Sensing **103**: 115-128.

Cabaneros, S. M., J. K. Calautit and B. R. Hughes (2019). "A review of artificial neural network models for ambient air pollution prediction." Environmental Modelling & Software **119**: 285-304.

Ebi, K. L., J. Vanos, J. W. Baldwin, J. E. Bell, D. M. Hondula, N. A. Errett, K. Hayes, C. E. Reid, S. Saha, J. Spector and P. Berry (2021). Extreme Weather and Climate Change: Population Health and Health System Implications. Annual Review of Public Health, Vol 42, 2021. J. E. Fielding. **42:** 293-315.

Escarda-Fernández, M., I. López-Riobóo-Botana, S. Barro-Tojeiro, L. Padrón-Cousillas, S. Gonzalez-Vázquez, A. Carreiro-Alonso and P. Gómez-Area (2024). "LLMs on the Fly: Text-to-JSON for Custom API Calling."

Frazier, A. E. and B. L. Hemingway (2021). "A Technical Review of Planet Smallsat Data: Practical Considerations for Processing and Using PlanetScope Imagery." Remote Sensing **13**(19).

Huang, J., P. Parthasarathi, M. Rezagholizadeh and S. Chandar (2024). "Towards Practical Tool Usage for Continually Learning LLMs." arXiv preprint arXiv:2404.09339.

Lairgi, Y., L. Moncla, R. Cazabet, K. Benabdeslem and P. Cleau (2024). "iText2KG: Incremental Knowledge Graphs Construction Using Large Language Models." Arxiv.

Lan, Y., S. He, K. Liu, X. Zeng, S. Liu and J. Zhao (2021). "Path-based knowledge reasoning with textual semantic information for medical knowledge graph completion." Bmc Medical Informatics and Decision Making **21**(SUPPL 9).

Liang, Y., C. Wu, T. Song, W. Wu, Y. Xia, Y. Liu, Y. Ou, S. Lu, L. Ji, S. Mao, Y. Wang, L. Shou, M. Gong and N. Duan (2024). "TaskMatrix.AI: Completing Tasks by Connecting Foundation Models with Millions of APIs." Intelligent Computing **3**: 0063.

Liu, Y., C. Pei, L. Xu, B. Chen, M. Sun, Z. Zhang, Y. Sun, S. Zhang, K. Wang, H. Zhang, J. Li, G. Xie, X. Wen, X. Nie, M. Ma and D. Pei (2023). OpsEval: A Comprehensive IT Operations Benchmark Suite for Large Language Models.

Oladeji, O., S. S. Mousavi and M. Roston (2023). "AI-driven E-Liability Knowledge Graphs: A Comprehensive Framework for Supply Chain Carbon Accounting and Emissions Liability Management." Arxiv.

OpenAi, J. Achiam, S. Adler, S. Agarwal, L. Ahmad, I. Akkaya, F. Aleman, D. Almeida, J. Altenschmidt, S. Altman, S. Anadkat, R. Avila, I. Babuschkin, S. Balaji, V. Balcom, P. Baltescu, H. Bao, M. Bavarian, J. Belgum and B. Zoph (2023). "GPT-4 Technical Report."

Pan, J. Z., S. Razniewski, J.-C. Kalo, S. Singhania, J. Chen, S. Dietze, H. Jabeen, J. Omeliyanenko, W. Zhang, M. Lissandrini, R. Biswas, G. d. Melo, A. Bonifati, E. Vakaj, M. Dragoni and D. Graux (2023). "Large Language Models and Knowledge Graphs: Opportunities and Challenges." Arxiv.

Patil, S. G., T. Zhang, X. Wang and J. E. Gonzalez (2023). "Gorilla: Large language model connected with massive apis." arXiv preprint arXiv:2305.15334.

Qin Yunlong, Wang Yingying, Zhang Bingsong and W. Fan (2020). "Research and Implementation of Provincial Meteorological BigData Service Platform Based on Extranet." Meteorological Science and Technology 48(06): 823-828+854.

Sallam, M. (2023). "ChatGPT Utility in Healthcare Education, Research, and Practice: Systematic Review on the Promising Perspectives and Valid Concerns." Healthcare 11(6).

Samsi, S., D. Zhao, J. McDonald, B. Li, A. Michaleas, M. Jones, W. Bergeron, J. Kepner, D. Tiwari, V. Gadepally and Ieee (2023). From Words to Watts: Benchmarking the Energy Costs of Large Language Model Inference. IEEE High Performance Extreme Computing Virtual Conference (HPEC), Electr Network.

Schick, T., J. Dwivedi-Yu, R. Dessí, R. Raileanu, M. Lomeli, E. Hambro, L. Zettlemoyer, N. Cancedda and T. Scialom (2024). Toolformer: language models can teach themselves to use tools. Proceedings of the 37th International Conference on Neural Information Processing Systems. New Orleans, LA, USA, Curran Associates Inc.: Article 2997.

Schneider, P., M. Klettner, K. Jokinen, E. Simperl and F. Matthes (2024). "Evaluating Large Language Models in Semantic Parsing for Conversational Question Answering over Knowledge Graphs." Arxiv.

Shen, Y., K. Song, X. Tan, D. Li, W. Lu and Y. Zhuang (2024). "Hugginggpt: Solving ai tasks with chatgpt and its friends in hugging face." Advances in Neural Information Processing Systems 36.

Sindall, R., T. Mecrow, A. C. Queiroga, C. Boyer, W. Koon and A. E. Peden (2022). "Drowning risk and climate change: a state-of-the-art review." Injury Prevention 28(2): 185-191.

Slater, L. J., L. Arnal, M.-A. Boucher, A. Y. Y. Chang, S. Moulds, C. Murphy, G. Nearing, G. Shalev, C. Shen, L. Speight, G. Villarini, R. L. Wilby, A. Wood and M. Zappa (2023). "Hybrid forecasting: blending climate predictions with AI models." Hydrology and Earth System Sciences 27(9): 1865-1889.

Soman, K., P. W. Rose, J. H. Morris, R. E. Akbas, B. Smith, B. Peetoom, C. Villouta-Reyes, G. Cerono, Y. Shi, A. Rizk-Jackson, S. Israni, C. A. Nelson, S. Huang and S. E. Baranzini (2024). "Biomedical knowledge graph-optimized prompt generation for large language models." Bioinformatics 40(9).

Sun, Q., Y. Luo, W. Zhang, S. Li, J. Li, K. Niu, X. Kong and W. Liu (2024). "Docs2KG: Unified Knowledge Graph Construction from Heterogeneous Documents Assisted by Large Language Models." Arxiv.

Sun, Y., K. Deng, K. Ren, J. Liu, C. Deng and Y. Jin (2024). "Deep learning in statistical downscaling for deriving high spatial resolution gridded meteorological data: A systematic review." Isprs Journal of Photogrammetry and Remote Sensing 208: 14-38.

Sun, Z., X. Zhou and G. Li (2023). "Learned Index: A Comprehensive Experimental Evaluation." Proc. VLDB Endow. 16(8): 1992–2004.

Tang, N., J. Fan, F. Li, J. Tu, X. Du, G. Li, S. Madden and M. Ouzzani (2021). "RPT: Relational Pre-trained Transformer Is Almost All You Need towards Democratizing Data Preparation." Proceedings of the Vldb Endowment 14(8): 1254-1261.

Thoppilan, R., D. De Freitas, J. Hall, N. M. Shazeer, A. Kulshreshtha, H.-T. Cheng, A. Jin, T. Bos, L. Baker, Y. Du, Y. Li, H. Lee, H. S. Zheng, A. Ghafouri, M. Menegali, Y. Huang, M. Krikun, D. Lepikhin, J. Qin, D. Chen, Y. Xu, Z. Chen, A. Roberts, M. Bosma, Y. Zhou, C.-C. Chang, I. A. Krivokon, W. J. Rusch, M. Pickett, K. S. Meier-Hellstern, M. R. Morris, T. Doshi, R. D. Santos, T. Duke, J. H. Søraker, B. Zevenbergen, V.

Prabhakaran, M. Díaz, B. Hutchinson, K. Olson, A. Molina, E. Hoffman-John, J. Lee, L. Aroyo, R. Rajakumar, A. Butryna, M. Lamm, V. O. Kuzmina, J. Fenton, A. Cohen, R. Bernstein, R. Kurzweil, B. Aguera-Arcas, C. Cui, M. R. Croak, E. H. Chi and Q. Le (2022). "LaMDA: Language Models for Dialog Applications." ArXiv **abs/2201.08239**.

Tufano, R., A. Mastropaolo, F. Pepe, O. Dabić, M. Di Penta and G. Bavota (2024). Unveiling ChatGPT's Usage in Open Source Projects: A Mining-based Study. 2024 IEEE/ACM 21st International Conference on Mining Software Repositories (MSR), IEEE.

Wang, J., M. Chen, B. Hu, D. Yang, Z. Liu, Y. Shen, P. Wei, Z. Zhang, J. Gu, J. Zhou, J. Z. Pan, W. Zhang and H. Chen (2024). "Learning to Plan for Retrieval-Augmented Large Language Models from Knowledge Graphs." Arxiv.

Wang, Y., Y. Kordi, S. Mishra, A. Liu, N. A. Smith, D. Khashabi and H. Hajishirzi (2022). "Self-instruct: Aligning language models with self-generated instructions." arXiv preprint arXiv:2212.10560.

Yadav, A. K. and S. S. Chandel (2014). "Solar radiation prediction using Artificial Neural Network techniques: A review." Renewable & Sustainable Energy Reviews **33**: 772-781.

Yasunaga, M., H. Ren, A. Bosselut, P. Liang, J. Leskovec and L. Assoc Computat (2021). QA-GNN: Reasoning with Language Models and Knowledge Graphs for Question Answering. Conference of the North-American-Chapter of the Association-for-Computational-Linguistics - Human Language Technologies (NAACL-HLT), Electr Network.

Yu, X., C. Chai, G. Li and J. Liu (2022). "Cost-Based or Learning-Based? A Hybrid Query Optimizer for Query Plan Selection." Proc. VLDB Endow. **15**(13): 3924–3936.

Zhang, K., H. Chen, L. Li and W. Wang (2023). "Syntax error-free and generalizable tool use for llms via finite-state decoding." arXiv preprint arXiv:2310.07075.

Zhao, F., L. Lim, I. Ahmad, D. Agrawal and A. El Abbadi (2023). "LLM-SQL-Solver: Can LLMs Determine SQL Equivalence?" ArXiv **abs/2312.10321**.

Zhao, X., X. Zhou and G. Li (2024). "Chat2Data: An Interactive Data Analysis System with RAG, Vector Databases and LLMs." Proc. VLDB Endow. **17**(12): 4481–4484.

Zhou, X., G. Li, J. Feng, L. Liu and W. Guo (2023). "Grep: A Graph Learning Based Database Partitioning System." Proc. ACM Manag. Data **1**(1): Article 94.

Zhou, X., Z. Sun and G. Li (2024). "DB-GPT: Large Language Model Meets Database." Data Science and Engineering **9**(1): 102-111.

Zhu, Y., W. Zhang, M. Chen, H. Chen, X. Cheng, W. Zhang, H. Chen and M. Assoc Comp (2022). DualDE: Dually Distilling Knowledge Graph Embedding for Faster and Cheaper Reasoning. 15th ACM International Conference on Web Search and Data Mining (WSDM), Electr Network.